\documentclass[letterpaper]{article} % DO NOT CHANGE THIS
\usepackage{times}  % DO NOT CHANGE THIS
\usepackage{helvet}  % DO NOT CHANGE THIS
\usepackage{courier}  % DO NOT CHANGE THIS
\usepackage[hyphens]{url}  % DO NOT CHANGE THIS
\usepackage{graphicx} % DO NOT CHANGE THIS
\usepackage{caption} % DO NOT CHANGE THIS AND DO NOT ADD ANY OPTIONS TO IT
\usepackage{natbib}  % DO NOT CHANGE THIS AND DO NOT ADD ANY OPTIONS TO IT
\usepackage{aaai2026}  % DO NOT CHANGE THIS
\pubnote{}
\copyrighttext{}
\usepackage{algorithm}
\usepackage{algorithmic}
\usepackage{booktabs}
\usepackage{xcolor}  % 임시로 생성
\usepackage{amsfonts}
\usepackage{amsmath}
\usepackage{multirow}
\usepackage{graphicx}
\usepackage{subcaption} % 이 줄을 preamble에 추가해야 합니다

\usepackage{newfloat}
\usepackage{listings}
\DeclareCaptionStyle{ruled}{labelfont=normalfont,labelsep=colon,strut=off} % DO NOT CHANGE THIS
\floatstyle{ruled}
\newfloat{listing}{tb}{lst}{}
\floatname{listing}{Listing}
\title{KeyRe-ID: Keypoint-Guided Person Re-Identification using Part-Aware Representation in Videos}

\author{
    Jinseong Kim\textsuperscript{\rm 1},
    Jeonghoon Song\textsuperscript{\rm 2},
    Gyeongseon Baek\textsuperscript{\rm 2},
    Byeongjoon Noh\textsuperscript{\rm 1,2}\thanks{Corresponding author.}\\
}
\affiliations{
    \textsuperscript{\rm 1}Department of AI and Big Data, Soonchunhyang University, 22 Soonchunhyang-ro, Asan, 31538, South Korea\\
    \textsuperscript{\rm 2}Department of Future Convergence Technology, Soonchunhyang University, 22 Soonchunhyang-ro, Asan, 31538, South Korea\\
}

\usepackage{bibentry}
\begin{document}

\maketitle

\begin{abstract}
We propose \textbf{KeyRe-ID}, a keypoint-guided video-based person re-identification framework consisting of global and local branches that leverage human keypoints for enhanced spatiotemporal representation learning. The global branch captures holistic identity semantics through Transformer-based temporal aggregation, while the local branch dynamically segments body regions based on keypoints to generate fine-grained, part-aware features. Extensive experiments on MARS and iLIDS-VID benchmarks demonstrate state-of-the-art performance, achieving 91.73\% mAP and 97.32\% Rank-1 accuracy on MARS, and 96.00\% Rank-1 and 100.0\% Rank-5 accuracy on iLIDS-VID. Implementation details are available at: \texttt{https://github.com/JinSeong0115/KeyRe-ID}.
\end{abstract}


\section{Introduction}
Person Re-Identification (Re-ID) is the task of identifying and retrieving the same individual across non-overlapping camera views, and it has become a fundamental computer vision technology in a variety of real-world applications, including intelligent video surveillance, pedestrian tracking, autonomous driving, and smart city infrastructure ~\cite{NING2024122419,xiao2024deep,dou2023identity,somers2024keypoint}. Depending on the input modality, Re-ID can be categorized into image-based and video-based approaches. While image-based Re-ID extracts features from static still images, video-based Re-ID leverages temporally continuous frame sequences—referred to as tracklets, which are short video clips of a single person—to incorporate temporal dependencies and motion cues. This temporal modeling enables more robust performance in challenging environments characterized by occlusions, pose variations, and illumination changes ~\cite{wu2024temporal,huang2025vills,zhang2024cross}.

Early video-based Re-ID approaches primarily relied on CNNs to extract frame-level features, followed by post-hoc temporal aggregation methods ~\cite{mclaughlin2016recurrent,zhou2017see,zhao2019attribute,yang2020spatial}. Recently, Transformer-based architectures have gained attention due to their ability to model long-range spatiotemporal dependencies via multi-head self-attention mechanisms while preserving high-resolution feature representations—an advantage over conventional CNNs \cite{zhang2021spatiotemporal, alsehaim22vidtransreid, tang2022multi}. Nonetheless, existing Transformer-based Re-ID frameworks commonly tokenize input frames using fixed-size patches or horizontal stripes, which limits their ability to adapt to fine-grained part-level motion and pose variations across frames.

To address these limitations, we propose \textbf{KeyRe-ID}, a keypoint-guided video person Re-ID framework designed to enhance part-level spatiotemporal representation learning. Instead of solely relying on patch divisions, our method incorporates keypoints—extracted from an external pose estimation model—to dynamically guide part segmentation on top of the patch-based tokenization. This dual-source approach enables the network to learn anatomically aligned part representations that more accurately capture body structure and motion patterns across frames. Each part-specific feature is temporally aggregated after undergoing clip-level perturbation to improve robustness under pose transitions.

In addition, a global representation is learned via a transformer-based global branch, following the architecture of VID-Trans-ReID \cite{alsehaim22vidtransreid}, enabling holistic clip-level identity modeling. The joint optimization of both global and keypoint-guided local branches enhances the model’s robustness to viewpoint changes, partial occlusions, and misalignment issues prevalent in real-world multi-camera environments. Extensive experiments on MARS and iLIDS-VID demonstrate that KeyRe-ID achieves state-of-the-art performance, validating its effectiveness in complex video-based Re-ID scenarios.

Extensive experiments conducted on two benchmark datasets, MARS \cite{zheng2016mars} and iLIDS-VID \cite{wang2014person}, validate the effectiveness of the proposed KeyRe-ID. The model achieves state-of-the-art (SOTA) performance, attaining a Rank-1 accuracy of 97.32\% and mAP of 91.73\% on MARS, and a Rank-5 accuracy of 100\% on iLIDS-VID. To the best of our knowledge, this is the first Transformer-based video Re-ID framework that jointly addresses dynamic keypoint-guided part segmentation and temporal modeling, making a significant step toward practical and robust deployment of Re-ID systems in complex real-world environments.

\section{Related Works}
Conventional video-based Re-ID approaches primarily rely on CNNs to extract frame-level visual features, followed by post-hoc temporal aggregation. However, these methods suffer from limited receptive fields and aggressive downsampling, which hinder their ability to capture spatiotemporal interactions effectively~\cite{kim2025spatio, he2021transreid}. To address this, Vision Transformer (ViT)-based architectures have been introduced, enabling direct modeling of long-range dependencies across frames~\cite{liu2023deeply}. Models such as VID-Trans-ReID~\cite{alsehaim22vidtransreid}, ST-Former~\cite{zhang2021spatiotemporal}, and Dual-Path Transformer~\cite{xia2024attention} leverage spatiotemporal attention over sampled clips to learn global representations~\cite{wu2024temporal}. Yet, most of these methods rely on fixed patch-level tokenization, making them less effective at capturing fine-grained part-level motion and pose variations~\cite{zhang2023completed}.

To overcome this, part-aware methods have incorporated horizontal stripes or semantic parsing to model local body regions~\cite{ren2022s, zhu2022learning, kalayeh2018human}. However, these approaches depend on static partitioning or pre-defined parsing maps, limiting their adaptability to dynamic body movements and temporal misalignments~\cite{wang2022pose, wei2021flexible}. While some works utilize keypoints or parsing masks for part-level cues, they often lack temporal consistency and are not integrated within a unified framework~\cite{he2023region, dou2022human, zhou2020fine}.

To address these gaps, our work builds on Transformer-based global modeling and introduces a unified framework that incorporates keypoint-guided dynamic part segmentation with temporal alignment. By jointly learning global and part-aware features, the proposed model enhances identification performance in complex, real-world scenarios.

\section{Proposed Method}
\subsection{Preliminary}
\begin{figure*}[!t]
  \centering
  \includegraphics[width=0.86\linewidth]{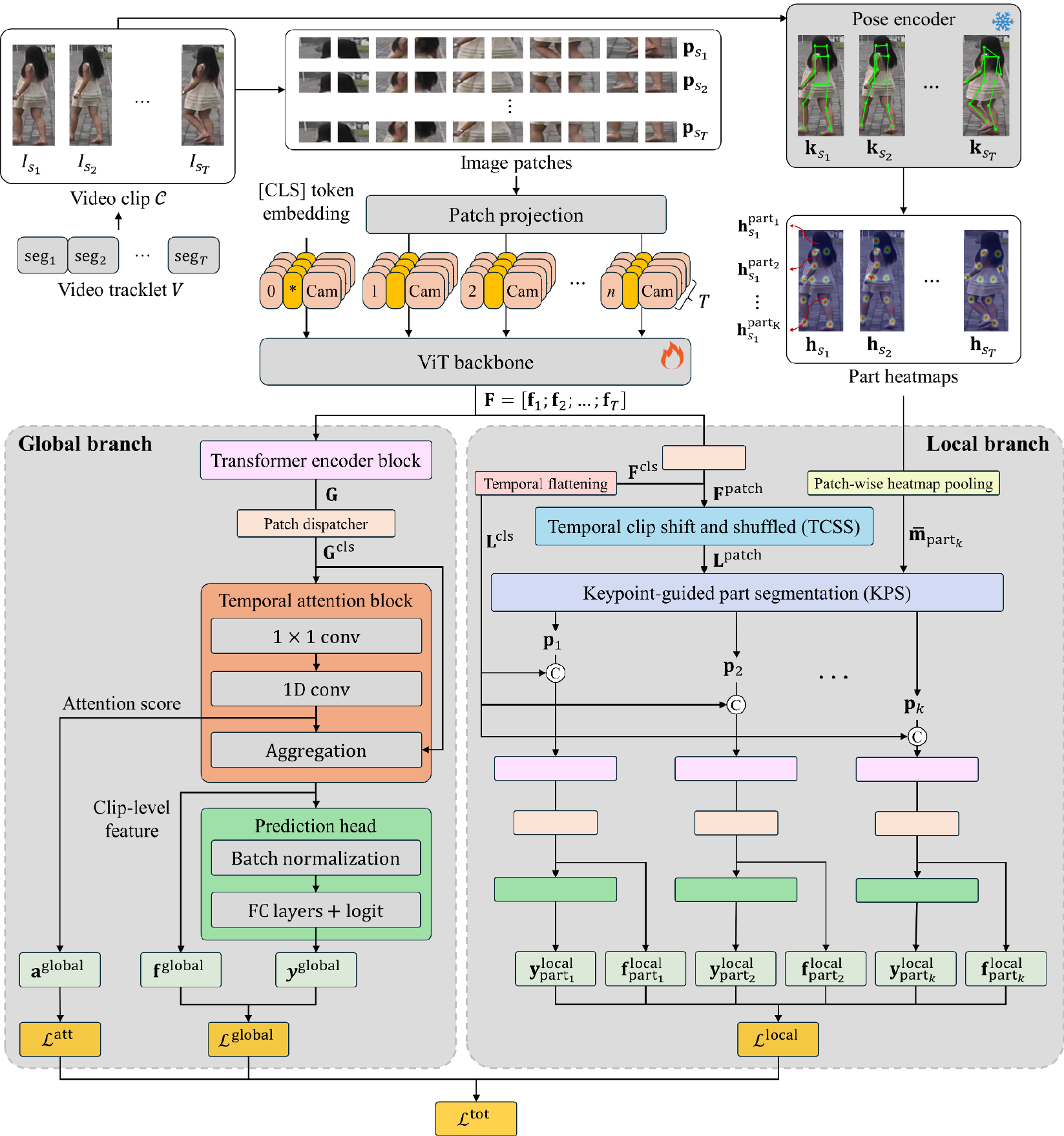}
  \caption{Overview of the training phase in the proposed \textbf{KeyRe-ID} framework. A video tracklet is segmented into temporal clips and processed by a shared \textit{ViT backbone} to extract deep spatiotemporal features. The framework consists of two branches: the \textit{Global branch} captures holistic identity cues by applying temporal attention over the extracted features, producing a clip-level representation. In parallel, the \textit{Local branch} first perturbs patch-wise representations using the \textit{TCSS module} to enhance temporal robustness. Subsequently, keypoint heatmaps are aggregated at the patch level and used in the \textit{KPS module} to generate body-part-specific features. Both branches are independently supervised by dedicated loss functions, and their outputs are jointly optimized through a composite objective to improve identity discrimination across pose and appearance variations.}
  \label{fig:framework}
\end{figure*}

In this section, we describe the proposed \textbf{KeyRe-ID} framework for video-based person re-identification. As illustrated in Figure~\ref{fig:framework}, the architecture comprises a shared spatiotemporal patch encoder followed by two specialized branches: a \textit{global branch} that captures holistic clip-level features and a \textit{local branch} that models fine-grained, part-aware cues guided by body keypoints. This design facilitates robust identity discrimination across varying poses and appearances. We begin by formalizing the input structure and intermediate representations that underpin our framework.

Given a video tracklet \( V = \{ I_1, I_2, \dots, I_N \} \) consisting of \(N\) RGB frames, we construct a temporally representative clip \( \mathcal{C} = \{ I_{s_1}, \dots, I_{s_T} \} \) by dividing the sequence into \(T\) equal segments and sampling one frame per segment. This promotes temporal diversity during training and continuity during inference.

\subsection{Video feature representation}
\subsubsection{ViT backbone}
The ViT backbone serves as the primary feature extractor, encoding spatial semantics from each video frame while preserving the overall temporal structure. Each input frame \(I_{s_t}\) is partitioned into overlapping patches of size \(P \times P\) with stride \(s\), yielding \(n\) patches per frame. Each patch is linearly projected into a \(D\)-dimensional embedding, and a learnable [CLS] token is prepended to each sequence to summarize global frame-level information.

For a sampled clip composed of \(T\) frames, the resulting token sequence is represented as:
\begin{equation}
\mathbf{Z} \in \mathbb{R}^{T \times (n+1) \times D}
\end{equation}
where each token group consists of \(n\) patch tokens and one [CLS] token.

To encode spatial and camera-related priors, we apply a learnable spatiotemporal positional embedding \(\mathbf{pos}\) and a camera-aware embedding \(\mathbf{cam}\), integrated as follows:
\begin{equation}
\tilde{\mathbf{Z}} = \mathbf{Z} + \lambda \cdot \mathbf{pos} + (1 - \lambda) \cdot \mathbf{cam}
\end{equation}

This enriched sequence \(\tilde{\mathbf{Z}}\) is then passed through the ViT backbone, producing spatiotemporal features, \(\mathbf{F} \in \mathbb{R}^{T \times (n+1) \times D}\)
which capture both local patch-level and global frame-level context. The output is decomposed into:
\begin{equation}
\mathbf{F}^{\text{cls}} \in \mathbb{R}^{T \times D}, \quad
\mathbf{F}^{\text{patch}} \in \mathbb{R}^{T \times n \times D}
\end{equation}
where \(\mathbf{F}^{\text{cls}}\) denotes the [CLS] token embeddings summarizing each frame, and \(\mathbf{F}^{\text{patch}}\) represents the embeddings of all patch tokens.

This architectural design allows the global and local branches to specialize in complementary identity cues—clip-level semantics and part-level appearance—contributing to robust person re-identification under diverse motion and viewpoint conditions.

\subsubsection{Keypoint-based part heatmap generation}
To improve part-aware representation learning in video-based person re-identification, this study utilizes keypoint heatmap information that reflects the human body structure. Prior video-based re-identification approaches have primarily relied on RGB visual cues, which often fail to robustly capture local identity-discriminative features under challenging conditions such as pose variations, occlusions, and background interference. To address these limitations, our framework incorporates structural spatial information derived from human joint locations. This design guides the model to attend to semantically meaningful body regions during feature learning, enabling the extraction of robust representations even under spatial misalignment and visual noise.

For each sampled frame \(I_{s_t}\), human keypoints are extracted using a pretrained PifPaf pose estimator~\cite{kreiss2019pifpaf}, which returns \(J\) predefined joint coordinates. Each joint location is then transformed into a 2D Gaussian heatmap aligned with the input frame resolution:
\begin{equation}
    \mathbf{h}_{s_t}^{j} \in \mathbb{R}^{H \times W}, \quad j = 1, \dots, J
\end{equation}
where \(\mathbf{h}_{s_t}^{j}\) represents the spatial likelihood distribution centered at the estimated joint position.

The extracted joints are then grouped into \(K\) semantically meaningful body parts, such as head, torso, arms, and legs, as summarized in Tab.~\ref{tab:parts-joints}.
% part-joint 표 추가 및 언급
Each part \(k\) is associated with a set of joints denoted by \(\mathcal{J}_k\), and the corresponding part-level heatmap is obtained by averaging the heatmaps of its constituent joints:
\begin{equation}
    \mathbf{h}_{s_t}^{\text{part}_k} = \frac{1}{|\mathcal{J}_k|} \sum_{j \in \mathcal{J}_k} \mathbf{h}_{s_t}^{j}, \quad k = 1, \dots, K
\end{equation}

These part-specific heatmaps preserve the structural spatial distribution of the person in each frame and serve as auxiliary information for local feature learning. In subsequent modules, they are aligned with visual patch tokens to facilitate region-level attention, thereby improving the discriminative power of local features under various spatial perturbations.

\begin{table}[!ht]
  \centering  
  \fontsize{9pt}{10pt}\selectfont
  \begin{tabular}{@{}ll@{}}
    \toprule
    \textbf{Part} & \textbf{Joints} \\
    \midrule
    Head & Eyes (left/right), ears (left/right), nose \\
    Torso & Shoulders (left/right), hip (left/right) \\
    Arms (left/right) & Elbows and wrists (left/right) \\
    Legs (left/right) & Knees and ankles (left/right) \\
    \bottomrule
  \end{tabular}
  \caption{Definition of the six body parts used in our framework, based on grouped keypoint joints.}
  \label{tab:parts-joints}
  \vspace{-0.3cm}
\end{table}

%%%%%%%%%%%%%%%%%%%%%%%%%%%%%%%%%%%%%%%%%%%%%%%%%%%%%%%%%%%%%%%%%%%%%%
\subsection{Global branch}
The primary role of the global branch is to predict the identity of the person depicted in the input video clip, denoted as \textit{Global\_ID}. To this end, it produces a discriminative clip-level representation by temporally aggregating frame-wise semantic features. Instead of explicitly modeling long-range temporal dependencies, this branch focuses on adaptive temporal integration using attention mechanisms. The overall design follows the principles of \textit{VID-Trans-ReID} model~\cite{alsehaim22vidtransreid}, while incorporating a transformer encoder for enhanced contextual reasoning.

Given the frame-wise feature sequence \(\mathbf{F} \in \mathbb{R}^{T \times (n+1) \times D}\) generated by the ViT backbone, we first pass it through a dedicated \textit{transformer encoder block} \(\mathcal{T}\), which enables spatiotemporal feature refinement via multi-head self-attention across all tokens:
\begin{equation}
    \mathbf{G} = \mathcal{T}(\mathbf{F}) \in \mathbb{R}^{T \times (n+1) \times D}
\end{equation}

To prepare the input for \textit{temporal attention block}, we extract the set of [CLS] tokens from \(\mathbf{G}\), denoted as:
\begin{equation}
    \mathbf{G}^{\text{cls}} = \{ \mathbf{g}_1^{\text{cls}}, \dots, \mathbf{g}_T^{\text{cls}} \} \in \mathbb{R}^{T \times D}
\end{equation}
These tokens compactly summarize frame-level semantics and serve as the foundation for frame-wise attention computation and subsequent clip-level representation learning. At this stage, a \textit{patch dispatcher} is used to decouple the [CLS] tokens from the remaining patch tokens. This separation enables the model to process global clip representations and local patch-wise representations independently in the subsequent branches.

\subsubsection{Temporal attention and aggregation}
To identify temporally salient frames that contribute most to identity discrimination, we apply a temporal attention module composed of a 2D convolution followed by a 1D temporal convolution applied to the frame-wise [CLS] features. This operation produces a raw attention score vector \(\mathbf{a}^{\text{global}} \in \mathbb{R}^{T}\). The scores are normalized using a softmax function to obtain temporal attention weights \(\boldsymbol{\alpha} = \{\alpha_1, \dots, \alpha_T\}\), which quantify the relative importance of each frame within the video clip.

Using these attention weights, we compute the clip-level representation as a weighted sum of the [CLS] tokens:
\begin{equation}
    \mathbf{f}^{\text{global}} = \sum_{t=1}^{T} \alpha_t \cdot \mathbf{g}^{\text{cls}}_t
\end{equation}

\begin{figure}[t]
  \centering
  \includegraphics[width=\linewidth]{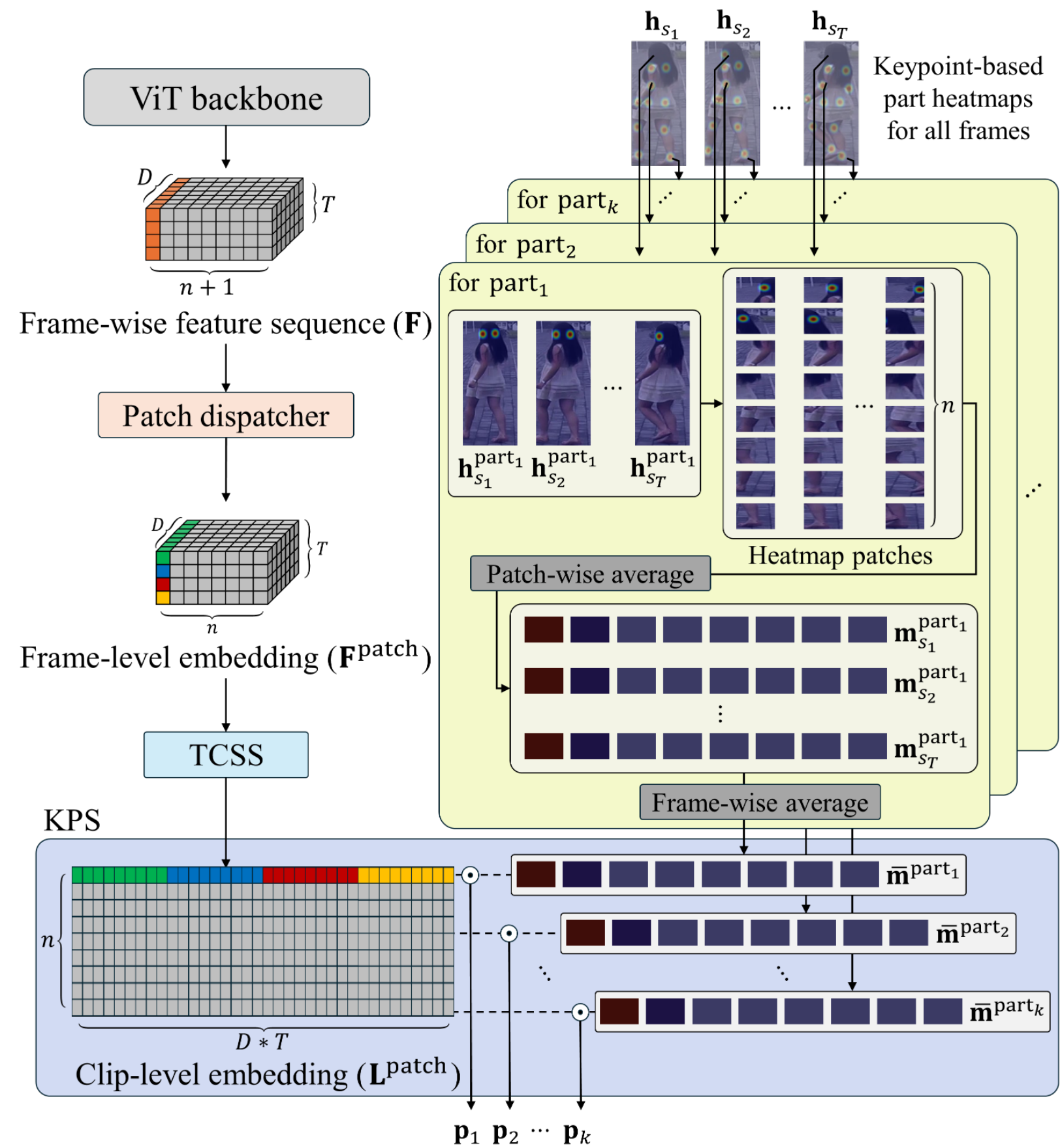}
  \caption{In the \textit{KPS module}, the operation to obtain part-aware spatio-temporal features involves averaging the part heatmaps across patch and frame dimensions to produce part-specific embeddings, which are then combined with the clip-level features from the \textit{TCSS} module through element-wise multiplication.}
  \label{fig:kps-framework}
  \vspace{-8pt}
\end{figure}

\subsubsection{Prediction head}
The aggregated feature \(\mathbf{f}^{\text{global}}\) is passed through a \textit{prediction head} consisting of batch normalization and fully connected layers, yielding the final global embedding \(\mathbf{y}^{\text{global}}\). This embedding is directly used to predict the clip-level identity label \textit{Global\_ID}, which corresponds to the person represented in the video.

All outputs from this branch—including the attention scores \(\mathbf{a}^{\text{global}}\), the clip-level representation \(\mathbf{f}^{\text{global}}\), and the final global embedding \(\mathbf{y}^{\text{global}}\)—are jointly optimized via a composite loss function. This function incorporates classification loss, triplet loss, and an attention regularization loss. Further implementation details are provided in the \textbf{Model optimization} section.

%%%%%%%%%%%%%%%%%%%%%%%%%%%%%%%%%%%%%%%%%%%%%%%%%%%%%%%%%%%%%%%%%%

\subsection{Local branch}
The local branch extracts fine-grained and part-aware discriminative features by integrating keypoint-guided spatial cues into the video-based Re-ID process. Unlike the global branch that captures holistic clip-level semantics, the local branch enhances robustness against occlusions and pose variations through temporal perturbations and body part-level aggregation.

\subsubsection{Temporal clip shift and shuffle (TCSS)}
Patch-level features in video-based Re-ID can suffer from overfitting to fixed temporal patterns or frame-specific biases, especially under challenging conditions like occlusion or abrupt motion. To mitigate this, we adopt the \textit{TCSS module}, which perturbs the temporal structure of patch tokens, promoting more robust feature learning.

Instead of relying on strict sequential order, \textit{TCSS module} applies a circular temporal shift to each patch-level feature across frames, followed by a patch-wise shuffling operation~\cite{zhang2018shufflenet}. This introduces temporal diversity, reduces reliance on exact frame order, and encourages richer feature representations. Our TCSS extends the original \textit{VID-Trans-ReID} design~\cite{alsehaim22vidtransreid} by subsequently incorporating keypoint-guided part information.

\subsubsection{Patch-wise heatmap pooling}
To guide patch-level representation learning, we construct part-specific patch importance vectors from detected human keypoints. The keypoints of each frame are grouped into \(K\) semantically meaningful body parts, and part heatmaps are obtained by averaging the corresponding joint heatmaps. These part heatmaps are downsampled via average pooling to match the patch grid, producing per-frame part-to-patch importance vectors.

For each part \(k\), we average these vectors over \(T\)  frames to obtain a clip-level importance vector \(\bar{\mathbf{m}}_{\text{part}_k}\). This vector semantically aligns body parts with patch tokens, facilitating robust feature encoding under pose and motion variations.

\subsubsection{Keypoint-guided part segmentation (KPS)}
As one of the core contributions of this study, the \textit{KPS module} addresses the limitations of conventional methods that rely solely on static patch structures, constructing semantically aligned local features by integrating keypoint-guided part importance with temporally structured patch representation.

As shown in Figure~\ref{fig:kps-framework}, \textit{KPS module} utilizes temporally ordered patch features \(\mathbf{L}^{\text{patch}}\in \mathbb{R}^{n \times (T \cdot D)}\), obtained through TCSS, and the clip-level part-specific importance vectors \(\bar{\mathbf{m}}^{\text{part}_k}\). These vectors encode patch relevance per body part, serving as soft attention maps. 

To obtain part-aware spatio-temporal features, we perform element-wise multiplication: \(\mathbf{p}_k = \mathbf{L}^{\text{patch}} \odot \bar{\mathbf{m}}_{\text{part}_k}^\top\). The result \(\mathbf{p}_k\) is concatenated with the [CLS] token embedding \(\mathbf{L}^{\text{cls}}\), which is the temporally flattened vector of \(\mathbf{F}^{\text{cls}}\), and passed through a transformer encoder for context enrichment.

The output sequence from the transformer encoder is subsequently passed through a patch dispatcher; the [CLS]-associated token is used as the part-level identity feature \(\mathbf{f}_{\text{part}_k}^\text{local}\), which is then passed through the \textit{prediction head} to produce the part-level identity feature \(\mathbf{y}_{\text{part}_k}^\text{local}\), following the same design as in the \textit{global branch}. The overall local supervision is applied through the local loss
\(\mathcal{L}^{\text{local}}\), which is jointly optimized with the global loss \(\mathcal{L}^{\text{global}}\) as part of the total training objective.

\subsection{Model optimization}\label{sec:optimization}
The proposed KeyRe-ID framework is trained in an end-to-end manner through two phases: training and inference.

\subsubsection{Training phase}
During training, the model is optimized with multi-branch loss strategy that jointly supervises global and part-aware (local) representations. The overall loss integrates global and local branch objectives as follows:
\begin{equation}
\mathcal{L}^{\text{tot}} = \alpha \cdot \mathcal{L}^{\text{global}} + (1-\alpha) \cdot \mathcal{L}^{\text{local}}
\end{equation}
where \(\alpha_g\) and \(\alpha_l\) are trade-off parameters. This formulation facilitates complementary learning between coarse-grained and fine-grained features.

The \textit{global branch} loss \(\mathcal{L}^{\text{global}}\) supervises a clip-level embedding that aggregates temporal identity information. It combines cross-entropy identity classification loss with label smoothing~\cite{szegedy2016rethinking}, triplet loss for inter-class margin maximization~\cite{hermans2017defense}, center loss for intra-class compactness~\cite{wen2016discriminative}, and attention regularization to penalize reliance on potentially occluded frames~\cite{pathak2020video}. Formally, it is defined as:
\begin{equation}
    \mathcal{L}^{\text{global}} = \mathcal{L}^{\text{ID}} + \mathcal{L}^{\text{triplet}} + \beta \cdot \mathcal{L}^{\text{center}} + \mathcal{L}^{\text{attn}}
\end{equation}
where \(\lambda_{\text{center}}\) is a weighting coefficients. The classification loss \(\mathcal{L}_{\text{ID}}\) is computed from the predicted identity \(\mathbf{y}^{\text{global}}\), while the triplet and center losses are applied to the embedding \(\mathbf{f}^{\text{global}}\).

The \textit{local branch} loss \(\mathcal{L}^{\text{local}}\) supervises part-specific embeddings, independently optimized for each body part. Similar to the \textit{global branch}, it includes identity classification, triplet loss, and center loss for each part \(k\): 
\begin{equation}
    \mathcal{L}^{\text{local}} = \frac{1}{K} \sum_{k=1}^{K} \left( \mathcal{L}_{k}^{\text{ID}} + \mathcal{L}_{k}^{\text{triplet}} + \beta \cdot \mathcal{L}_{k}^{\text{center}} \right)
\end{equation}
This enables fine-grained supervision of local features, enhancing robustness to occlusions and pose variations. All model components are jointly optimized through back-propagation.

\subsubsection{Inference phase}
During inference, the KeyRe-ID model identifies individuals by comparing the extracted features of a query tracklet with those in a gallery, which stores concatenated global and part-aware embeddings of previously observed persons, typically collected from multi-camera environments. For each query, the model extracts a global embedding and multiple part-specific embeddings, corresponding to predefined body regions. These features are concatenated to form the final descriptor, which is compared against all gallery entries using Euclidean distance.

\begin{table*}[!ht]
  \centering
  \fontsize{9pt}{10pt}\selectfont
  \setlength{\tabcolsep}{4pt}
  \resizebox{\textwidth}{!}{%
  \begin{tabular}{@{} l|cc|cc | l|cc|cc @{}}
    \toprule
    \multirow{2}{*}[-0.5ex]{\textbf{Method}} & \multicolumn{2}{c|}{\textbf{MARS}} & \multicolumn{2}{c|}{\textbf{iLIDS-VID}} &
    \multirow{2}{*}[-0.5ex]{\textbf{Method}} & \multicolumn{2}{c|}{\textbf{MARS}} & \multicolumn{2}{c}{\textbf{iLIDS-VID}} \\
    \cmidrule{2-5} \cmidrule{7-10}
     & \textbf{mAP} & \textbf{Rank-1} & \textbf{Rank-1} & \textbf{Rank-5} &
     & \textbf{mAP} & \textbf{Rank-1} & \textbf{Rank-1} & \textbf{Rank-5} \\
    \midrule

    STMP~\cite{liu2019spatial} & 72.7 & 84.4 & 84.3 & 96.8            & CTL~\cite{liu2021spatial} & 86.7 & 91.4 & 89.7 & 97.0 \\ 
    M3D~\cite{li2020multi} & 74.1 & 84.4 & 74.0 & 94.3                & PSTA~\cite{wang2021pyramid} & 85.8 & 91.5 & 91.5 & 98.1 \\ 
    GLTR~\cite{li2019global} & 78.5 & 87.0 & 86.0 & 98.0              & KSTL~\cite{guo2023key} & 86.3 & 91.5 & 93.4 & – \\    
    RethinkingTF~\cite{jiang2020rethinking} & 85.2 & 87.1 & 87.7 & –  & CRF~\cite{bai2024incorporating} & 87.4 & 91.5 & 93.4 & – \\
    Co-Aware~\cite{wu2021person} & 84.1 & 88.2 & 85.8 & –             & TCViT~\cite{wu2024temporal} & 87.6 & 91.7 & 94.3 & 99.3 \\       
    VRSTC~\cite{hou2019vrstc} & 82.3 & 88.5 & 83.4 & 95.5             & KMPN (MGH)~\cite{chen2022keypoint} & 86.6 & 92.0 & – & – \\        
    MG-RAFA~\cite{zhang2020multi} & 85.6 & 88.8 & 88.6 & 98.0         & PSFormer~\cite{zhu2025not} & 88.2 & 92.1 & 93.7 & 99.5  \\   
    MGH~\cite{yan2020learning} & 85.8 & 90.0 & 85.6 & 97.1            & DCCT~\cite{liu2023deeply} & 87.5 & 92.3 & 91.7 & 98.6 \\       
    SSN3D~\cite{jiang2021ssn3d} & 86.2 & 90.1 & – & –                 & TF-CLIP~\cite{yu2024tf} & 89.4 & 93.0 & 94.5 & 99.1 \\
    PiT~\cite{zang2022multidirection} & 86.8 & 90.2 & 92.1 & 98.9     & CLIMB-ReID~\cite{yu2025climb} & 89.7 & 93.3 & \textbf{96.7} & 99.9 \\
    GRL~\cite{liu2021watching} & 84.8 & 91.0 & 90.4 & 98.3            & STFE~\cite{yang2024stfe} & 91.5 & 95.5 & – & – \\
    SINet~\cite{bai2022salient} & 86.2 & 91.0 & 92.5 & –              & VID-Trans-ReID~\cite{alsehaim22vidtransreid} & 90.2 & 96.36 & 94.67 & – \\
    DSANet~\cite{kim2023feature} & 86.6 & 91.1 & 85.1 & 75.5          & \textbf{KeyRe-ID (Ours)} & \textbf{91.73} & \textbf{97.32} & 96.0 & \textbf{100.0} \\  
    \bottomrule
  \end{tabular}
  }

  \caption{Comparison of video-based person re-identification methods on the MARS and iLIDS-VID datasets, reporting mAP and Rank-1 for MARS, and Rank-1 and Rank-5 for iLIDS-VID (sorted by MARS Rank-1 ascending).}
  \vspace{-5pt}
  \label{tab:reid-comparison}
\end{table*}

\begin{figure*}[!hbt]
  \centering
  \begin{subfigure}{0.48\linewidth}
    \centering
    \includegraphics[width=\linewidth]{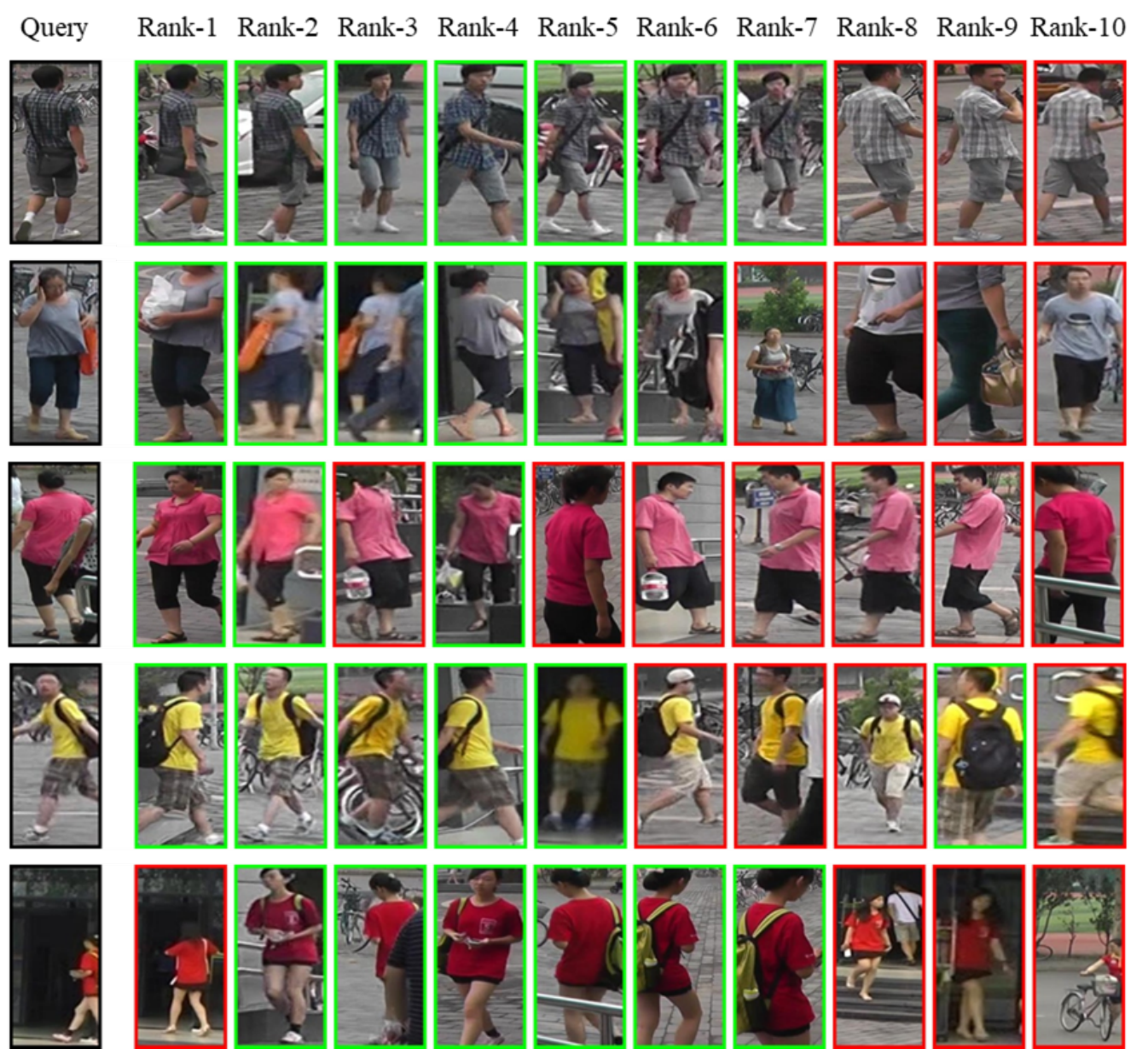}
    \caption{Retrieval results of \textit{VID-Trans-ReID}}
  \end{subfigure}
  \hfill
  \begin{subfigure}{0.48\linewidth}
    \centering
    \includegraphics[width=\linewidth]{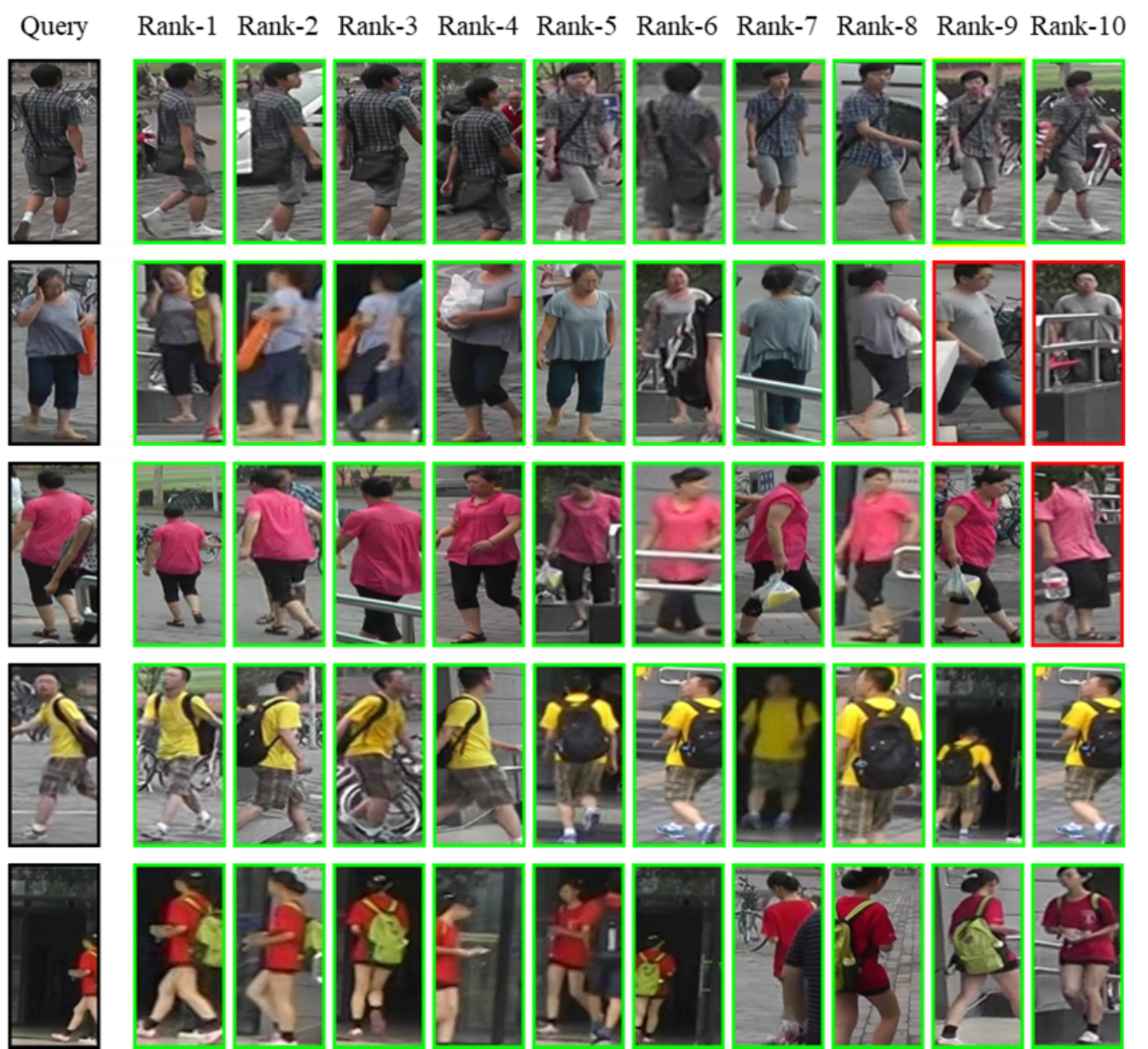}
    \caption{Retrieval results of \textit{KeyRe-ID}}
  \end{subfigure}
  \caption{Comparison of the results of top-10 retrieval results for two video-based Re-ID models on the MARS dataset. For each query image (leftmost), the top-10 ranked gallery images are shown from left to right. Green boxes denote correct matches, and red boxes denote incorrect ones.}
  \label{fig:ranking-list}
  \vspace{-5pt}
\end{figure*}

\section{Validation}
\subsection{Experimental setup}
\subsubsection{Datasets}
We evaluate our method on two widely used video-based person Re-ID benchmarks: 
\begin{itemize}
    \item \textbf{MARS}~\cite{zheng2016mars}: is the largest benchmark in this domain, comprising 1,261 identities and 20,715 tracklets captured from 6 cameras. 
    \item \textbf{iLIDS-VID}~\cite{Li_2018_ECCV}: contains 300 identities and 600 tracklets recorded across 2 cameras, offering more challenging variations in viewpoint and occlusion under limited data.
\end{itemize}

\begin{table}[!ht]
  \centering
  \fontsize{9pt}{10pt}\selectfont
  \setlength{\tabcolsep}{4pt}
  \begin{tabular}{@{} c c c c | c c @{}}
    \toprule
    \multicolumn{6}{c}{\textbf{MARS}} \\
    \midrule
    \textbf{Global branch} & \textbf{Local branch} & \textbf{TCSS} & \textbf{KPS} & \textbf{mAP} & \textbf{Rank-1} \\
    \midrule
    \checkmark & \checkmark & \(\times\) & \checkmark & 90.73 & 96.91 \\
    \checkmark & \checkmark & \checkmark & \(\times\) & 90.25 & 96.36 \\
    \checkmark & \(\times\) & \(\times\) & \(\times\) & 88.35 & 95.50 \\
    \(\times\) & \checkmark & \checkmark & \checkmark & 85.74 & 95.88 \\
    \midrule
    \checkmark & \checkmark & \checkmark & \checkmark & \textbf{91.73} & \textbf{97.32} \\
    \toprule
    \multicolumn{6}{c}{\textbf{iLIDS-VID}} \\
    \midrule 
    \textbf{Global branch} & \textbf{Local branch} & \textbf{TCSS} & \textbf{KPS} & \textbf{Rank-1} & \textbf{Rank-5} \\
    \midrule
    \checkmark & \checkmark & \(\times\) & \checkmark & 94.00 & 100.00 \\
    \checkmark & \checkmark & \checkmark & \(\times\) & 94.67 & 99.50 \\
    \checkmark & \(\times\) & \(\times\) & \(\times\) & 89.33 & 98.67 \\
    \(\times\) & \checkmark & \checkmark & \checkmark & 94.67 & 97.33 \\
    \midrule
    \checkmark & \checkmark & \checkmark & \checkmark & \textbf{96.00} & \textbf{100.00} \\
    \bottomrule
  \end{tabular}
  \caption{Ablation study results under different architectural configurations on the MARS and iLIDS‐VID datasets. The check mark (\checkmark) indicates that the corresponding component is included, 
    while the cross (\(\times\)) denotes its exclusion.}
  \vspace{-5pt}
  \label{tab:ablation-study}
\end{table}

\begin{figure}[t]
  \centering

  \begin{subfigure}{0.4\textwidth}
    \centering
    \includegraphics[width=\linewidth]{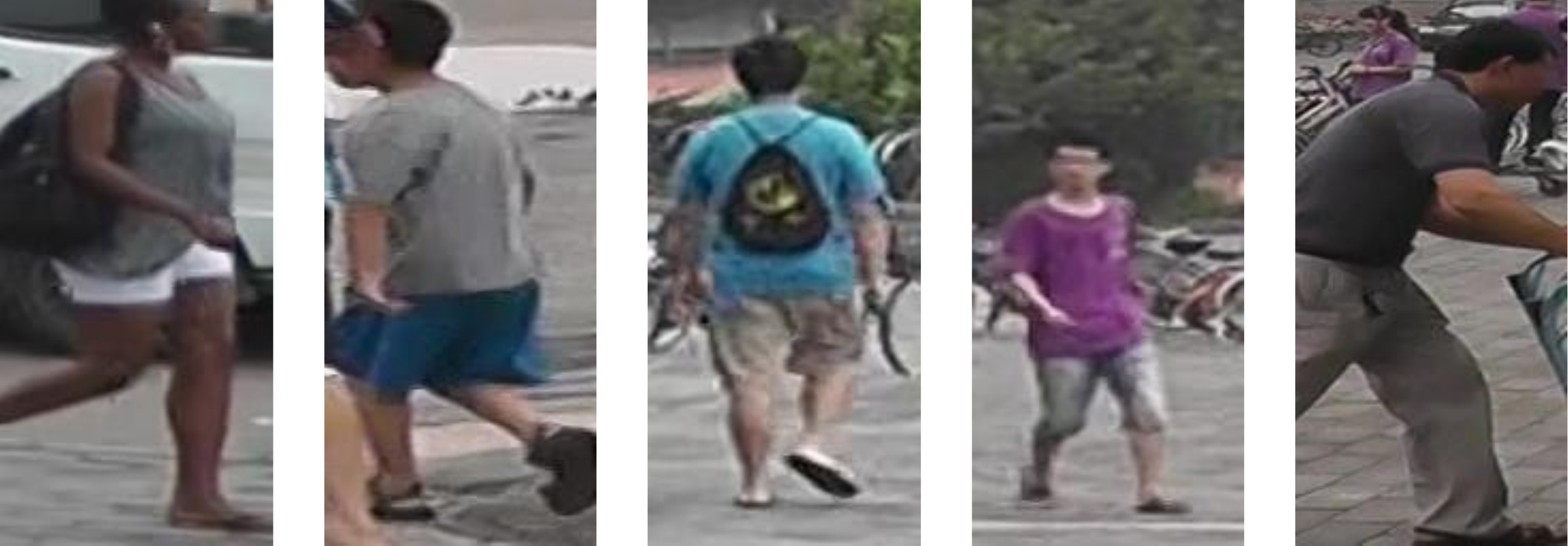}
    \caption{}
  \end{subfigure}
  \vspace{1mm}

  \begin{subfigure}{0.4\textwidth}
    \centering
    \includegraphics[width=\linewidth]{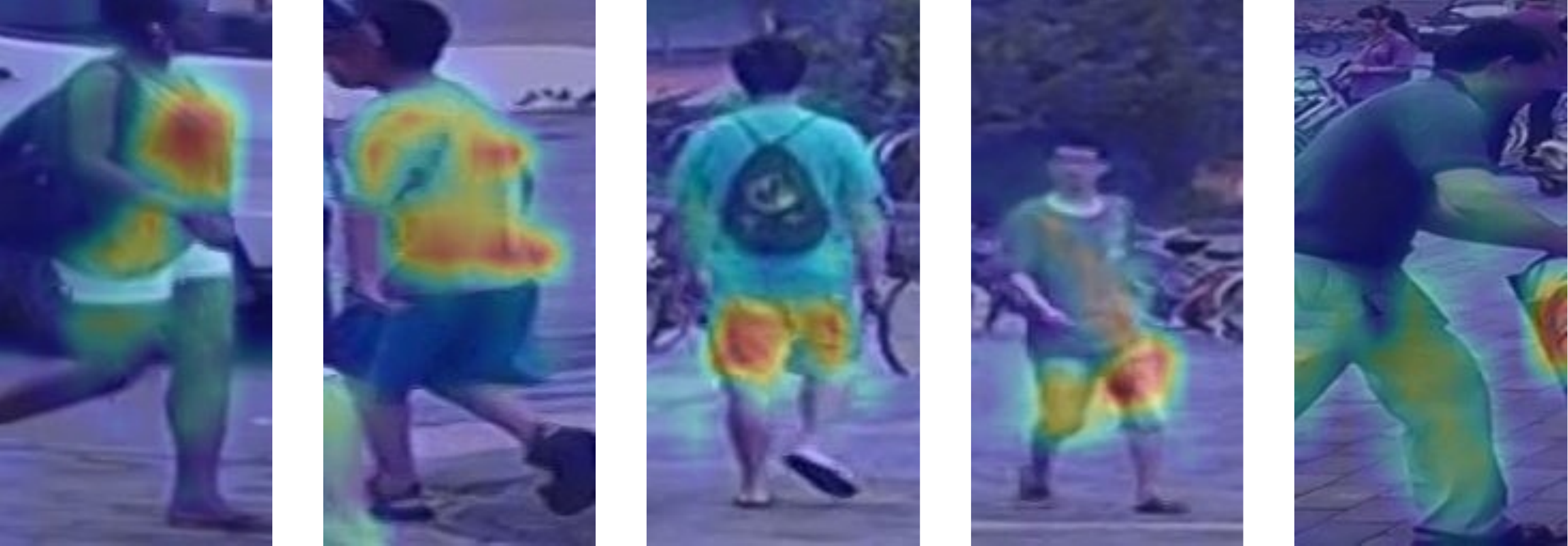}
    \caption{}
  \end{subfigure}
  \vspace{1mm}

  \begin{subfigure}{0.4\textwidth}
    \centering
    \includegraphics[width=\linewidth]{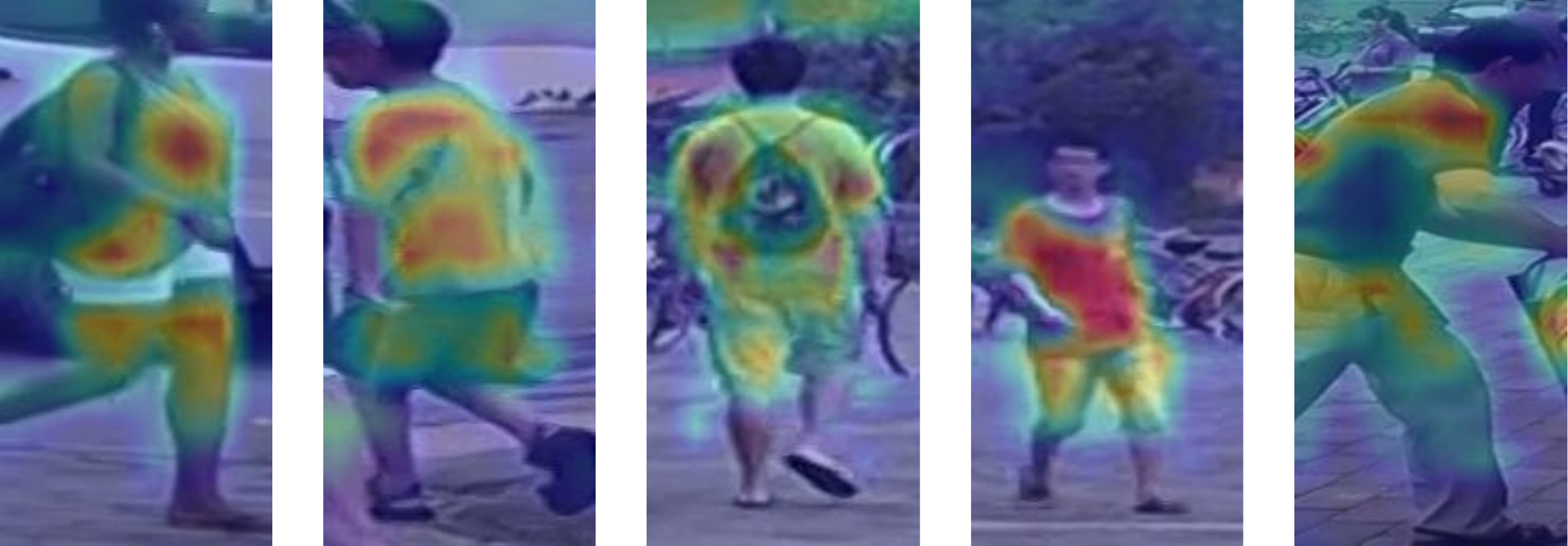}
    \caption{}
  \end{subfigure}

    \caption{Attention maps on (a) input query images: (b) by VID-Trans-ReID that exhibits scattered and context-insensitive focus across the body, and (c) by KeyRe-ID that shows more consistent and semantically meaningful activations on identity-discriminative regions.}

  \label{fig:attention-map}
  \vspace{-8pt}
\end{figure}

\subsubsection{Evaluation metrics}
On MARS, we report mean Average Precision (mAP) and Rank-1 accuracy due to its multi-shot structure. For iLIDS-VID, we report Rank-1 and Rank-5 accuracy. mAP measures average retrieval precision across ranks, while Rank-k indicates the probability of finding a correct match within the top-k retrieved results.

\subsubsection{Baselines}
We compare our method with a wide range of Re-ID models. Among them, \textit{VID-Trans-ReID}~\cite{alsehaim22vidtransreid}, which achieves the highest Rank-1 accuracy on the MARS dataset, is selected as our primary comparison target.

\subsubsection{Implementation details}
Input frames are resized to \(256 \times 128\), and the model is trained for 120 epochs using SGD (learning rate: 0.008, batch size: 32, momentum: 0.9)) with cosine scheduling and 5 warm-up epochs. Each training sample is a randomly augmented 4-frame clip with horizontal flipping and random erasing. We adopt a 12-layer ViT backbone with number of patches (\(P=16\), stride \(s=16\), dimension \(D = 768\)) pretrained on ImageNet-21K. TCSS shift size is set to 5, and KPS uses 6 parts from 17 keypoints. Loss weights are \(\alpha=0.75\) and \(\beta=0.0005\).

\subsection{Main results}
Tab.~\ref{tab:reid-comparison} presents the results of the proposed KeyRe-ID method compared to other methods on MARS and iLIDS-VID datasets. On MARS, the proposed KeyRe-ID achieves SOTA performance with an mAP of 91.73\% and Rank-1 accuracy of 97.32\%, surpassing \textit{VID-Trans-ReID} model (mAP 90.25\%, Rank‐1 96.36\%). On iLIDS-VID, our method attains Rank-1 accuracy of 96.00\% and perfect Rank-5 accuracy (100.0\%), surpassing the \textit{CLIMB-ReID}'s Rank-5 (99.9\%) accuracy, and closely matching its Rank-1 (96.7\%). These results confirm that the proposed method effectively combines transformer-based global modeling with keypoint-guided part-aware learning via TCSS and KPS, leading to improved spatiotemporal identity discrimination and setting new benchmarks for video person Re-ID.

\subsubsection{Ranking list}
Figure~\ref{fig:ranking-list} shows qualitative top-10 retrieval results on the MARS dataset. VID-Trans-ReID frequently yields incorrect matches (red boxes) due to the inherent limitations of stripe-based segmentation, especially under challenging conditions involving pose variations and similar local appearances across different individuals. Conversely, KeyRe-ID, benefiting from KPS-based semantic segmentation, consistently delivers accurate top-ranking matches (green boxes). This clearly illustrates the robustness and improved discriminative capability of KeyRe-ID, attributable to anatomically-aligned, part-aware feature representations.

\subsubsection{Attention map comparison}
We provide qualitative comparisons between VID-Trans-ReID and KeyRe-ID through attention map visualizations (Figure~\ref{fig:attention-map}). VID-Trans-ReID, which relies on fixed stripe-based segmentation, produces rigid and narrowly localized activations, limiting its adaptability to diverse human poses. In contrast, KeyRe-ID integrates keypoint-guided spatial priors to dynamically align attention maps with semantically meaningful body parts. This structured attention not only enables broader and more balanced spatial coverage, but also enhances interpretability by clearly highlighting informative regions of each body part. As a result, KeyRe-ID effectively captures fine-grained identity cues, contributing to improved retrieval accuracy and robustness under varied visual conditions.

\subsection{Ablation studies}
To assess the contribution of each component in the proposed framework, we conduct comprehensive ablation experiments, as shown in Tab.~\ref{tab:ablation-study}. Removing the TCSS module results in a noticeable decline in performance, indicating its importance in enhancing temporal robustness against frame-level misalignment and motion perturbation. Excluding the KPS module reduces the mAP by 1.48\% on the MARS dataset, highlighting the role of keypoint-guided spatial guidance in improving part-aware feature discrimination. The removal of the local branch leads to a significant performance drop, especially on MARS, underscoring its critical role in capturing fine-grained identity cues. Similarly, removing the global branch causes a marked degradation in overall accuracy (MARS mAP: 85.74\%, Rank-1: 95.88\%), demonstrating the necessity of holistic feature modeling. Collectively, the full KeyRe-ID model, incorporating all proposed components, achieves the highest performance across datasets, validating the complementary strengths and synergistic benefits of joint global and local representation learning.

\section{Conclusion}
In this paper, we proposed KeyRe-ID, a novel video-based person re-identification framework that leverages human keypoints to guide part-level representation learning. Built upon ViT backbone, KeyRe-ID integrates global and local branches, with the global branch aggregating clip-level semantic features via temporal attention, and the local branch employing KPS to construct fine-grained, part-aware features robust to pose variations. Extensive experiments on two widely-used benchmarks, MARS and iLIDS-VID, demonstrated the effectiveness of our method, achieving SOTA performance with Rank-1 accuracy of 97.32\% and mAP of 91.73\% on MARS, as well as Rank-1 accuracy of 96.00\% and Rank-5 accuracy of about 100.0\% on iLIDS-VID. Qualitative analysis confirmed that the KPS module enhances feature alignment with human anatomical structures, thereby improving the accuracy of top-ranked retrieval results. Ablation studies validated the contributions of both TCSS and KPS modules in boosting overall performance.

% \section{Acknowledgments}

% \bibliographystyle{named}  % 이 줄 추가
% \bibliography{aaai2026}

\begin{thebibliography}{58}
\providecommand{\natexlab}[1]{#1}

\bibitem[{Alsehaim and Breckon(2022)}]{alsehaim22vidtransreid}
Alsehaim, A.; and Breckon, T. 2022.
\newblock VID-Trans-ReID: Enhanced Video Transformers for Person Re-identification.
\newblock In \emph{Proc. British Machine Vision Conference}. BMVA.

\bibitem[{Bai, Chang, and Ma(2024)}]{bai2024incorporating}
Bai, S.; Chang, H.; and Ma, B. 2024.
\newblock Incorporating texture and silhouette for video-based person re-identification.
\newblock \emph{Pattern Recognition}, 156: 110759.

\bibitem[{Bai et~al.(2022)Bai, Ma, Chang, Huang, and Chen}]{bai2022salient}
Bai, S.; Ma, B.; Chang, H.; Huang, R.; and Chen, X. 2022.
\newblock Salient-to-broad transition for video person re-identification.
\newblock In \emph{Proceedings of the IEEE/CVF conference on computer vision and pattern recognition}, 7339--7348.

\bibitem[{Chen et~al.(2022)Chen, Doering, Zhang, Yang, Gall, and Schiele}]{chen2022keypoint}
Chen, D.; Doering, A.; Zhang, S.; Yang, J.; Gall, J.; and Schiele, B. 2022.
\newblock Keypoint message passing for video-based person re-identification.
\newblock In \emph{Proceedings of the AAAI Conference on Artificial Intelligence}, volume~36, 239--247.

\bibitem[{Dou et~al.(2022)Dou, Zhao, Jiang, Zhang, Zheng, and Zuo}]{dou2022human}
Dou, S.; Zhao, C.; Jiang, X.; Zhang, S.; Zheng, W.-S.; and Zuo, W. 2022.
\newblock Human co-parsing guided alignment for occluded person re-identification.
\newblock \emph{IEEE Transactions on Image Processing}, 32: 458--470.

\bibitem[{Dou et~al.(2023)Dou, Wang, Li, and Wang}]{dou2023identity}
Dou, Z.; Wang, Z.; Li, Y.; and Wang, S. 2023.
\newblock Identity-seeking self-supervised representation learning for generalizable person re-identification.
\newblock In \emph{Proceedings of the IEEE/CVF international conference on computer vision}, 15847--15858.

\bibitem[{Guo and Wang(2023)}]{guo2023key}
Guo, W.; and Wang, H. 2023.
\newblock Key Parts Spatio-Temporal Learning for Video Person Re-identification.
\newblock In \emph{Proceedings of the 5th ACM International Conference on Multimedia in Asia}, 1--6.

\bibitem[{He et~al.(2023)He, Chen, Wang, Luo, Wang, Jiang, and Ding}]{he2023region}
He, S.; Chen, W.; Wang, K.; Luo, H.; Wang, F.; Jiang, W.; and Ding, H. 2023.
\newblock Region generation and assessment network for occluded person re-identification.
\newblock \emph{IEEE transactions on information forensics and security}, 19: 120--132.

\bibitem[{He et~al.(2021)He, Luo, Wang, Wang, Li, and Jiang}]{he2021transreid}
He, S.; Luo, H.; Wang, P.; Wang, F.; Li, H.; and Jiang, W. 2021.
\newblock Transreid: Transformer-based object re-identification.
\newblock In \emph{Proceedings of the IEEE/CVF international conference on computer vision}, 15013--15022.

\bibitem[{Hermans, Beyer, and Leibe(2017)}]{hermans2017defense}
Hermans, A.; Beyer, L.; and Leibe, B. 2017.
\newblock In defense of the triplet loss for person re-identification.
\newblock \emph{arXiv preprint arXiv:1703.07737}.

\bibitem[{Hou et~al.(2019)Hou, Ma, Chang, Gu, Shan, and Chen}]{hou2019vrstc}
Hou, R.; Ma, B.; Chang, H.; Gu, X.; Shan, S.; and Chen, X. 2019.
\newblock Vrstc: Occlusion-free video person re-identification.
\newblock In \emph{Proceedings of the IEEE/CVF conference on computer vision and pattern recognition}, 7183--7192.

\bibitem[{Huang et~al.(2025)Huang, Prabhakar, Guo, Chellappa, and Peng}]{huang2025vills}
Huang, S.; Prabhakar, R.; Guo, Y.; Chellappa, R.; and Peng, C. 2025.
\newblock VILLS: Video-Image Learning to Learn Semantics for Person Re-Identification.
\newblock In \emph{2025 IEEE/CVF Winter Conference on Applications of Computer Vision (WACV)}, 5969--5979. IEEE.

\bibitem[{Jiang et~al.(2020)Jiang, Gong, Guo, Yang, Huang, Zheng, Zheng, and Sun}]{jiang2020rethinking}
Jiang, X.; Gong, Y.; Guo, X.; Yang, Q.; Huang, F.; Zheng, W.-S.; Zheng, F.; and Sun, X. 2020.
\newblock Rethinking temporal fusion for video-based person re-identification on semantic and time aspect.
\newblock In \emph{Proceedings of the AAAI Conference on Artificial Intelligence}, volume~34, 11133--11140.

\bibitem[{Jiang et~al.(2021)Jiang, Qiao, Yan, Li, Zheng, and Chen}]{jiang2021ssn3d}
Jiang, X.; Qiao, Y.; Yan, J.; Li, Q.; Zheng, W.; and Chen, D. 2021.
\newblock SSN3D: Self-separated network to align parts for 3D convolution in video person re-identification.
\newblock In \emph{Proceedings of the AAAI conference on artificial intelligence}, volume~35, 1691--1699.

\bibitem[{Kalayeh et~al.(2018)Kalayeh, Basaran, G{\"o}kmen, Kamasak, and Shah}]{kalayeh2018human}
Kalayeh, M.~M.; Basaran, E.; G{\"o}kmen, M.; Kamasak, M.~E.; and Shah, M. 2018.
\newblock Human semantic parsing for person re-identification.
\newblock In \emph{Proceedings of the IEEE conference on computer vision and pattern recognition}, 1062--1071.

\bibitem[{Kim et~al.(2025)Kim, Cho, Lee, and Lee}]{kim2025spatio}
Kim, M.; Cho, M.; Lee, H.; and Lee, S. 2025.
\newblock Spatio-temporal Feature-level Augmentation Vision Transformer for video-based person re-identification.
\newblock \emph{Pattern Recognition}, 111813.

\bibitem[{Kim, Cho, and Lee(2023)}]{kim2023feature}
Kim, M.; Cho, M.; and Lee, S. 2023.
\newblock Feature disentanglement learning with switching and aggregation for video-based person re-identification.
\newblock In \emph{Proceedings of the IEEE/CVF Winter Conference on Applications of Computer Vision}, 1603--1612.

\bibitem[{Kreiss, Bertoni, and Alahi(2019)}]{kreiss2019pifpaf}
Kreiss, S.; Bertoni, L.; and Alahi, A. 2019.
\newblock Pifpaf: Composite fields for human pose estimation.
\newblock In \emph{Proceedings of the IEEE/CVF conference on computer vision and pattern recognition}, 11977--11986.

\bibitem[{Li et~al.(2019)Li, Wang, Tian, Gao, and Zhang}]{li2019global}
Li, J.; Wang, J.; Tian, Q.; Gao, W.; and Zhang, S. 2019.
\newblock Global-local temporal representations for video person re-identification.
\newblock In \emph{Proceedings of the IEEE/CVF international conference on computer vision}, 3958--3967.

\bibitem[{Li, Zhang, and Huang(2020)}]{li2020multi}
Li, J.; Zhang, S.; and Huang, T. 2020.
\newblock Multi-scale temporal cues learning for video person re-identification.
\newblock \emph{IEEE Transactions on Image Processing}, 29: 4461--4473.

\bibitem[{Li, Zhu, and Gong(2018)}]{Li_2018_ECCV}
Li, M.; Zhu, X.; and Gong, S. 2018.
\newblock Unsupervised Person Re-identification by Deep Learning Tracklet Association.
\newblock In \emph{Proceedings of the European Conference on Computer Vision (ECCV)}.

\bibitem[{Liu et~al.(2019)Liu, Sun, Xu, Xu, and Yu}]{liu2019spatial}
Liu, J.; Sun, C.; Xu, X.; Xu, B.; and Yu, S. 2019.
\newblock A spatial and temporal features mixture model with body parts for video-based person re-identification.
\newblock \emph{Applied Intelligence}, 49: 3436--3446.

\bibitem[{Liu et~al.(2021{\natexlab{a}})Liu, Zha, Wu, Zheng, and Sun}]{liu2021spatial}
Liu, J.; Zha, Z.-J.; Wu, W.; Zheng, K.; and Sun, Q. 2021{\natexlab{a}}.
\newblock Spatial-temporal correlation and topology learning for person re-identification in videos.
\newblock In \emph{Proceedings of the IEEE/CVF conference on computer vision and pattern recognition}, 4370--4379.

\bibitem[{Liu et~al.(2023)Liu, Yu, Zhang, and Lu}]{liu2023deeply}
Liu, X.; Yu, C.; Zhang, P.; and Lu, H. 2023.
\newblock Deeply coupled convolution--transformer with spatial--temporal complementary learning for video-based person re-identification.
\newblock \emph{IEEE Transactions on Neural Networks and Learning Systems}.

\bibitem[{Liu et~al.(2021{\natexlab{b}})Liu, Zhang, Yu, Lu, and Yang}]{liu2021watching}
Liu, X.; Zhang, P.; Yu, C.; Lu, H.; and Yang, X. 2021{\natexlab{b}}.
\newblock Watching you: Global-guided reciprocal learning for video-based person re-identification.
\newblock In \emph{Proceedings of the IEEE/CVF conference on computer vision and pattern recognition}, 13334--13343.

\bibitem[{McLaughlin, Del~Rincon, and Miller(2016)}]{mclaughlin2016recurrent}
McLaughlin, N.; Del~Rincon, J.~M.; and Miller, P. 2016.
\newblock Recurrent convolutional network for video-based person re-identification.
\newblock In \emph{Proceedings of the IEEE conference on computer vision and pattern recognition}, 1325--1334.

\bibitem[{Ning et~al.(2024)Ning, Wang, Zhang, Ning, and Tiwari}]{NING2024122419}
Ning, E.; Wang, C.; Zhang, H.; Ning, X.; and Tiwari, P. 2024.
\newblock Occluded person re-identification with deep learning: A survey and perspectives.
\newblock \emph{Expert Systems with Applications}, 239: 122419.

\bibitem[{Pathak, Eshratifar, and Gormish(2020)}]{pathak2020video}
Pathak, P.; Eshratifar, A.~E.; and Gormish, M. 2020.
\newblock Video person re-id: Fantastic techniques and where to find them (student abstract).
\newblock In \emph{Proceedings of the AAAI Conference on Artificial Intelligence}, volume~34, 13893--13894.

\bibitem[{Ren et~al.(2022)Ren, Zhang, Bao, and Zhang}]{ren2022s}
Ren, X.; Zhang, D.; Bao, X.; and Zhang, Y. 2022.
\newblock S\textsuperscript{2}-Net: Semantic and Saliency Attention Network for Person Re-Identification.
\newblock \emph{IEEE Transactions on Multimedia}, 25: 4387--4399.

\bibitem[{Somers, Alahi, and Vleeschouwer(2024)}]{somers2024keypoint}
Somers, V.; Alahi, A.; and Vleeschouwer, C.~D. 2024.
\newblock Keypoint promptable re-identification.
\newblock In \emph{European Conference on Computer Vision}, 216--233. Springer.

\bibitem[{Szegedy et~al.(2016)Szegedy, Vanhoucke, Ioffe, Shlens, and Wojna}]{szegedy2016rethinking}
Szegedy, C.; Vanhoucke, V.; Ioffe, S.; Shlens, J.; and Wojna, Z. 2016.
\newblock Rethinking the inception architecture for computer vision.
\newblock In \emph{Proceedings of the IEEE conference on computer vision and pattern recognition}, 2818--2826.

\bibitem[{Tang et~al.(2022)Tang, Zhang, Peng, Chen, and Lin}]{tang2022multi}
Tang, Z.; Zhang, R.; Peng, Z.; Chen, J.; and Lin, L. 2022.
\newblock Multi-stage spatio-temporal aggregation transformer for video person re-identification.
\newblock \emph{IEEE Transactions on Multimedia}, 25: 7917--7929.

\bibitem[{Wang et~al.(2014)Wang, Gong, Zhu, and Wang}]{wang2014person}
Wang, T.; Gong, S.; Zhu, X.; and Wang, S. 2014.
\newblock Person re-identification by video ranking.
\newblock In \emph{Computer Vision--ECCV 2014: 13th European Conference, Zurich, Switzerland, September 6-12, 2014, Proceedings, Part IV 13}, 688--703. Springer.

\bibitem[{Wang et~al.(2022)Wang, Liu, Song, Guo, and Shi}]{wang2022pose}
Wang, T.; Liu, H.; Song, P.; Guo, T.; and Shi, W. 2022.
\newblock Pose-guided feature disentangling for occluded person re-identification based on transformer.
\newblock In \emph{Proceedings of the AAAI conference on artificial intelligence}, volume~36, 2540--2549.

\bibitem[{Wang et~al.(2021)Wang, Zhang, Gao, Geng, Lu, and Wang}]{wang2021pyramid}
Wang, Y.; Zhang, P.; Gao, S.; Geng, X.; Lu, H.; and Wang, D. 2021.
\newblock Pyramid spatial-temporal aggregation for video-based person re-identification.
\newblock In \emph{Proceedings of the IEEE/CVF international conference on computer vision}, 12026--12035.

\bibitem[{Wei et~al.(2021)Wei, Yang, Wang, and Gao}]{wei2021flexible}
Wei, Z.; Yang, X.; Wang, N.; and Gao, X. 2021.
\newblock Flexible body partition-based adversarial learning for visible infrared person re-identification.
\newblock \emph{IEEE Transactions on Neural Networks and Learning Systems}, 33(9): 4676--4687.

\bibitem[{Wen et~al.(2016)Wen, Zhang, Li, and Qiao}]{wen2016discriminative}
Wen, Y.; Zhang, K.; Li, Z.; and Qiao, Y. 2016.
\newblock A discriminative feature learning approach for deep face recognition.
\newblock In \emph{Computer vision--ECCV 2016: 14th European conference, amsterdam, the netherlands, October 11--14, 2016, proceedings, part VII 14}, 499--515. Springer.

\bibitem[{Wu et~al.(2021)Wu, Ye, Lin, Gao, and Shen}]{wu2021person}
Wu, D.; Ye, M.; Lin, G.; Gao, X.; and Shen, J. 2021.
\newblock Person re-identification by context-aware part attention and multi-head collaborative learning.
\newblock \emph{IEEE transactions on information forensics and security}, 17: 115--126.

\bibitem[{Wu et~al.(2024)Wu, Wang, Zhou, Hua, and Sun}]{wu2024temporal}
Wu, P.; Wang, L.; Zhou, S.; Hua, G.; and Sun, C. 2024.
\newblock Temporal correlation vision transformer for video person re-identification.
\newblock In \emph{Proceedings of the AAAI Conference on Artificial Intelligence}, volume~38, 6083--6091.

\bibitem[{Xia et~al.(2024)Xia, Tan, Dai, Zhao, Wu, and Cao}]{xia2024attention}
Xia, J.; Tan, L.; Dai, P.; Zhao, M.; Wu, Y.; and Cao, L. 2024.
\newblock Attention disturbance and dual-path constraint network for occluded person re-identification.
\newblock In \emph{Proceedings of the AAAI Conference on Artificial Intelligence}, volume~38, 6198--6206.

\bibitem[{Xiao et~al.(2024)Xiao, Zhang, Ning, Lu, Song, Qing, and Chen}]{xiao2024deep}
Xiao, Z.; Zhang, Z.; Ning, Y.; Lu, Y.; Song, W.; Qing, X.; and Chen, J. 2024.
\newblock Deep learning in person re-identification: a survey.
\newblock In \emph{International Conference on Image Processing and Artificial Intelligence (ICIPAl 2024)}, volume 13213, 939--948. SPIE.

\bibitem[{Yan et~al.(2020)Yan, Qin, Chen, Liu, Zhu, Tai, and Shao}]{yan2020learning}
Yan, Y.; Qin, J.; Chen, J.; Liu, L.; Zhu, F.; Tai, Y.; and Shao, L. 2020.
\newblock Learning multi-granular hypergraphs for video-based person re-identification.
\newblock In \emph{Proceedings of the IEEE/CVF conference on computer vision and pattern recognition}, 2899--2908.

\bibitem[{Yang et~al.(2020)Yang, Zheng, Yang, Chen, and Tian}]{yang2020spatial}
Yang, J.; Zheng, W.-S.; Yang, Q.; Chen, Y.-C.; and Tian, Q. 2020.
\newblock Spatial-temporal graph convolutional network for video-based person re-identification.
\newblock In \emph{Proceedings of the IEEE/CVF conference on computer vision and pattern recognition}, 3289--3299.

\bibitem[{Yang et~al.(2024)Yang, Wang, Liu, Wang, and Gao}]{yang2024stfe}
Yang, X.; Wang, X.; Liu, L.; Wang, N.; and Gao, X. 2024.
\newblock STFE: a comprehensive video-based person re-identification network based on spatio-temporal feature enhancement.
\newblock \emph{IEEE Transactions on Multimedia}, 26: 7237--7249.

\bibitem[{Yu et~al.(2024)Yu, Liu, Wang, Zhang, and Lu}]{yu2024tf}
Yu, C.; Liu, X.; Wang, Y.; Zhang, P.; and Lu, H. 2024.
\newblock TF-CLIP: Learning text-free CLIP for video-based person re-identification.
\newblock In \emph{Proceedings of the AAAI conference on artificial intelligence}, volume~38, 6764--6772.

\bibitem[{Yu et~al.(2025)Yu, Liu, Zhu, Wang, Zhang, and Lu}]{yu2025climb}
Yu, C.; Liu, X.; Zhu, J.; Wang, Y.; Zhang, P.; and Lu, H. 2025.
\newblock CLIMB-ReID: A Hybrid CLIP-Mamba Framework for Person Re-Identification.
\newblock In \emph{Proceedings of the AAAI Conference on Artificial Intelligence}, volume~39, 9589--9597.

\bibitem[{Zang, Li, and Gao(2022)}]{zang2022multidirection}
Zang, X.; Li, G.; and Gao, W. 2022.
\newblock Multidirection and multiscale pyramid in transformer for video-based pedestrian retrieval.
\newblock \emph{IEEE Transactions on Industrial Informatics}, 18(12): 8776--8785.

\bibitem[{Zhang et~al.(2024)Zhang, Luo, Cheng, Yang, Ran, Xing, and Zhang}]{zhang2024cross}
Zhang, S.; Luo, W.; Cheng, D.; Yang, Q.; Ran, L.; Xing, Y.; and Zhang, Y. 2024.
\newblock Cross-platform video person reid: A new benchmark dataset and adaptation approach.
\newblock In \emph{European Conference on Computer Vision}, 270--287. Springer.

\bibitem[{Zhang et~al.(2021)Zhang, Wei, Xie, Zhuang, Zhang, Li, and Tian}]{zhang2021spatiotemporal}
Zhang, T.; Wei, L.; Xie, L.; Zhuang, Z.; Zhang, Y.; Li, B.; and Tian, Q. 2021.
\newblock Spatiotemporal transformer for video-based person re-identification.
\newblock \emph{arXiv preprint arXiv:2103.16469}.

\bibitem[{Zhang et~al.(2018)Zhang, Zhou, Lin, and Sun}]{zhang2018shufflenet}
Zhang, X.; Zhou, X.; Lin, M.; and Sun, J. 2018.
\newblock Shufflenet: An extremely efficient convolutional neural network for mobile devices.
\newblock In \emph{Proceedings of the IEEE conference on computer vision and pattern recognition}, 6848--6856.

\bibitem[{Zhang et~al.(2023)Zhang, He, Liu, Xiao, and Durrani}]{zhang2023completed}
Zhang, Z.; He, D.; Liu, S.; Xiao, B.; and Durrani, T.~S. 2023.
\newblock Completed part transformer for person re-identification.
\newblock \emph{IEEE Transactions on Multimedia}, 26: 2303--2313.

\bibitem[{Zhang et~al.(2020)Zhang, Lan, Zeng, and Chen}]{zhang2020multi}
Zhang, Z.; Lan, C.; Zeng, W.; and Chen, Z. 2020.
\newblock Multi-granularity reference-aided attentive feature aggregation for video-based person re-identification.
\newblock In \emph{Proceedings of the IEEE/CVF conference on computer vision and pattern recognition}, 10407--10416.

\bibitem[{Zhao et~al.(2019)Zhao, Shen, Jin, Lu, and Hua}]{zhao2019attribute}
Zhao, Y.; Shen, X.; Jin, Z.; Lu, H.; and Hua, X.-s. 2019.
\newblock Attribute-driven feature disentangling and temporal aggregation for video person re-identification.
\newblock In \emph{Proceedings of the IEEE/CVF conference on computer vision and pattern recognition}, 4913--4922.

\bibitem[{Zheng et~al.(2016)Zheng, Bie, Sun, Wang, Su, Wang, and Tian}]{zheng2016mars}
Zheng, L.; Bie, Z.; Sun, Y.; Wang, J.; Su, C.; Wang, S.; and Tian, Q. 2016.
\newblock Mars: A video benchmark for large-scale person re-identification.
\newblock In \emph{Computer Vision--ECCV 2016: 14th European Conference, Amsterdam, The Netherlands, October 11-14, 2016, Proceedings, Part VI 14}, 868--884. Springer.

\bibitem[{Zhou et~al.(2020)Zhou, Zhong, Lan, Sun, Zhang, Zhang, and Ji}]{zhou2020fine}
Zhou, Q.; Zhong, B.; Lan, X.; Sun, G.; Zhang, Y.; Zhang, B.; and Ji, R. 2020.
\newblock Fine-grained spatial alignment model for person re-identification with focal triplet loss.
\newblock \emph{IEEE Transactions on Image Processing}, 29: 7578--7589.

\bibitem[{Zhou et~al.(2017)Zhou, Huang, Wang, Wang, and Tan}]{zhou2017see}
Zhou, Z.; Huang, Y.; Wang, W.; Wang, L.; and Tan, T. 2017.
\newblock See the forest for the trees: Joint spatial and temporal recurrent neural networks for video-based person re-identification.
\newblock In \emph{Proceedings of the IEEE conference on computer vision and pattern recognition}, 4747--4756.

\bibitem[{Zhu et~al.(2022)Zhu, Guo, Liu, Wang, and Tang}]{zhu2022learning}
Zhu, K.; Guo, H.; Liu, S.; Wang, J.; and Tang, M. 2022.
\newblock Learning semantics-consistent stripes with self-refinement for person re-identification.
\newblock \emph{IEEE transactions on neural networks and learning systems}.

\bibitem[{Zhu et~al.(2025)Zhu, Chen, Ji, Ye, and Liu}]{zhu2025not}
Zhu, L.; Chen, T.; Ji, D.; Ye, J.; and Liu, J. 2025.
\newblock Not Every Patch is Needed: Towards a More Efficient and Effective Backbone for Video-based Person Re-identification.
\newblock \emph{IEEE Transactions on Image Processing}.

\end{thebibliography}

  % <-- bbl 직접 포함

\end{document}